\documentclass{article}
\usepackage{arxiv}
\usepackage{lipsum}
\PassOptionsToPackage{fontenc}{T1}
\input{curvenote.def}

\usepackage{fontenc}    

\usepackage{xcolor}
\hypersetup{colorlinks = true,
linkcolor = purple,
urlcolor  = blue,
citecolor = cyan,
anchorcolor = black}

\title{HRPlanes: High Resolution Airplane Dataset for Deep Learning}

\date{}

\makeatletter
\let\@fnsymbol\@arabic
\makeatother

\usepackage{authblk}

\author[1]{Tolga Bakırman}
\author[2]{Elif Sertel}
\affil[1]{Yildiz Technical University}
\affil[2]{Istanbul Technical University}

\renewcommand{\shorttitle}{HRPlanes: High Resolution Airplane Dataset}

\hypersetup{
pdftitle={\@title},
pdfsubject={},
pdfauthor={\@author},
pdfkeywords={},
addtopdfcreator={Written in Curvenote}
}

\begin{document}
\maketitle

\begin{abstract}

Airplane detection from satellite imagery is a challenging task due to the complex backgrounds in the images and differences in data acquisition conditions caused by the sensor geometry and atmospheric effects. Deep learning methods provide reliable and accurate solutions for automatic detection of airplanes; however, huge amount of training data is required to obtain promising results. In this study, we create a novel airplane detection dataset called High Resolution Planes (HRPlanes) by using images from Google Earth (GE) and labeling the bounding box of each plane on the images. HRPlanes include GE images of several different airports across the world to represent a variety of landscape, seasonal and satellite geometry conditions obtained from different satellites. We evaluated our dataset with two widely used object detection methods namely YOLOv4 and Faster R-CNN. Our preliminary results show that the proposed dataset can be a valuable data source and benchmark data set for future applications. Moreover, proposed architectures and results of this study could be used for transfer learning of different datasets and models for airplane detection.

\end{abstract}


\section{Introduction}\label{a16601c4}

The rapid technological advancements in remote sensing systems has significantly improved the availability of very high resolution remote sensing imagery to be used for the detection of geospatial objects such as airplanes, ships, buildings, etc. \citep{LI2020296}. Airplane detection is essential in various fields such as airport surveillance, transportation activity analysis, defense and military applications and satellite imagery is a significant data source for this purpose with the advantages of covering large areas very quickly and periodically \citep{rs12030458}.

Airplane detection studies from earlier years are generally based on template matching and machine learning. For example, \citep{6353895} and \citep{XU20101759} have utilized deformable templates for airplane detection. Although this method is flexible and outperforms rigid shape matching, it still needs various types of information for template design \citep{CHENG201611}. Compared to template matching, machine learning methods have been used more widely for this purpose. Various feature extraction methods and classifiers are investigated in the literature. \citep{5982082} utilized a spatial sparse coding bag of words (BOW) model combined with linear support vector machine. This models uses sliding windows to extract features and employs spatial mapping strategy to encode geometric information. \citep{6512596} proposed rotation invariant histogram of oriented gradient (HOG) features for detection of complex objects in high resolution imagery. They also improve their method further by using generic discriminative part-based model later on \citep{ZHANG201530}. \citep{6043875} proposed a novel color-enhanced rotation-invariant Hough forest to train a Pose-Estimation-based Rotation-invariant Texton Forest. \citep{LIU20145327} investigated the airplane feature possessing rotation invariant that combined with sparse coding and radial gradient transform. Machine learning methods require manually extracted features and thus, their performance are heavily depend on selecting accurate hand-crafted features \citep{Ball2017Comprehensive}. Deep learning approaches offer end-to-end solutions using automatic feature extraction.

Recent studies illustrate that deep learning based airplane detection methods do not only outperform conventional object detection algorithms but also provide feasible solutions. \citep{rs10010139} combined classification and localization processes for better optimization using transfer learning. \citep{s18072335} proposed a multilayer feature fusion process that fuses the shallow and deep layer features in fully convolutional neural networks (FCN). \citep{rs11091062} utilized the L2 norm normalization, feature connection, scale scaling, and feature dimension reduction for more efficient fusion of low and high level features. \citep{rs12030458} assessed different deep learning approaches namely Faster Regional Convolutional Neural Network (Faster R-CNN), Single Shot Multi-box Detector (SSD) and You Only Look Once Version 3 (YOLOv3) for airplane detection from very high resolution satellite imagery. \citep{9178761} proposed Weakly Supervised Learning in AlexNet which requires only image-level labelled data contrary to other object detection methods. \citep{zhou2021aircraft} introduced Multiscale Detection Network to detect small scale aircrafts in a multiscale detection architecture manner. \citep{rs13112207} developed a Faster R-CNN based model that combines multi-angle features driven and majority voting strategy. \citep{Shi2021Aircraft} introduced Deconvolution operation with Position Attention mechanism that captures the external structural feature representation of aircraft during the feature map generation process. \citep{s21082618} proposed a self-calibrated Mask R-CNN model that performs perform object recognition and segmentation in parallel. \citep{Zeng2022Top} utilizes a top-down approach for aircraft detection in large scenes. Once the airport area is extracted with U-Net, Faster-RCNN with feature enhancement module is applied for target detection. \citep{s22010319} have proposed two stage aircraft detection network. The first stage creates region proposal using a circular intensity filter and the second stage detects targets by using combination of rotation-invariant histogram of oriented gradient and vector of locally aggregated descriptors.

In order to train deep neural networks, huge amounts of images and corresponding labels are needed. Researchers have proposed datasets including airplanes for this purpose. \citep{Xia2017DOTA} introduced DOTA dataset with 15 classes including airplanes using imagery from Google Earth, Jilin-1 and Gaofen-2 satellites. This dataset have been then expanded, improved and renamed as iSAID dataset \citep{Zamir2019iSAID}. \citep{Lam2018xView} utilized multi-source imagery to generate xView dataset which has passenger/cargo plane and 59 other classes. More recently, \citep{Shermeyer2020RarePlanes} took the advantage of synthetic data to create RarePlanes dataset. The dataset consists of 253 WorldView-3 real and 50,000 synthetic imageries, and corresponding 14,700 hand annotated and 630,000 simulated plane labels, respectively.

In this study, we create a huge novel dataset solely for airplanes from Google Earth imagery using only hand annotated labels. Our dataset, HRPlanes, include images obtained from the biggest airports across the world to represent a variety of landscape, seasonal and data acquisition geometry conditions. We also evaluate our dataset using state-of-the-art deep neural networks namely YOLOv4 and Faster R-CNN to analyze the performance of two different object detection algorithms.


\section{Dataset}\label{ad15d6e7}

The imagery required for the dataset has been obtained from Google Earth. We have downloaded 4800 x 2703 sized 3092 RGB images from the biggest airports of the world such as Paris-Charles de Gaulle, John F. Kennedy, Frankfurt, Istanbul, Madrid, Dallas, Las Vegas and Amsterdam Airports and aircraft boneyards like Davis-Monthan Air Force Base.

\begin{figure}[!htbp]
  \centering
  \includegraphics[width=1\linewidth]{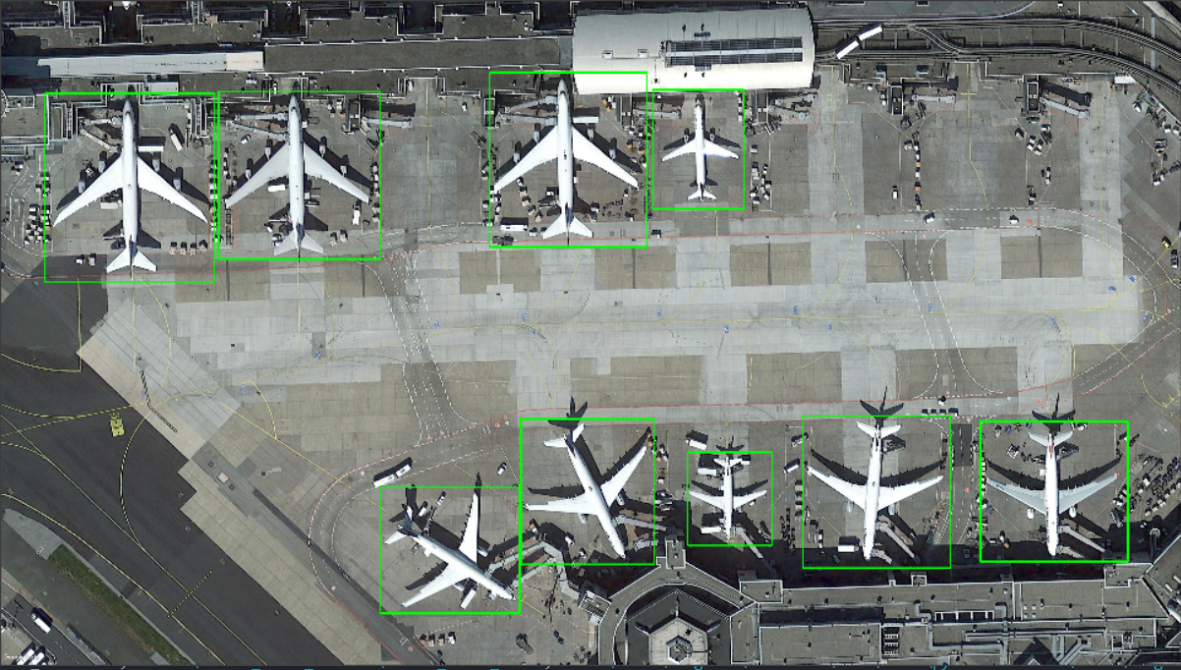}

  \caption{
    A sample image from the dataset
  }
  \label{cbyzYz838r}
\end{figure}

Dataset images were annotated manually by creating bounding boxes for each airplane using HyperLabel \citep{hyperlabel_2020} software. Quality control of each label was conducted by visual inspection of independent analysts who were not included in the labelling procedure. A total of 18,477 airplanes have been labelled. A sample image and corresponding minimum boxes for airplanes can be seen in Figure~\ref{cbyzYz838r}.

The dataset has been approximately split as 70\% (2166 images), 20\% (615 images) and 10\% (311 images) for training, validation and testing, respectively.

\section{Methods}\label{a42c9a34}

The proposed dataset has been evaluated using state-of-the-art YOLOv4 and Faster R-CNN methods. Both methods have been widely utilized in various object detection applications in the literature.

\subsection{YOLOv4}\label{abd61599}

The first version of YOLO is proposed by \citep{Redmon2015You} which is highlighted to be extremely fast and treating detection problem as regression. The network which runs on Darknet framework is a global model since it uses features from the whole image to predict bounding boxes. Therefore, the model is able learn general representations of the objects. The input image is divided by \textit{S x S} grid and the grid that covers the center of an object is also to detect it. The bounding boxes are predicted with a confidence score which is calculated by: \textit{Pr\textsubscript{(object)} x IoU\textsubscript{GT}}. YOLO has a GoogLeNet \citep{7298594} based architecture which has 24 convolutional and 2 fully connected layers. The network does not utilize inception module, instead it uses \textit{1 x 1} reduction layer with \textit{3 x 3} convolutional layer. While linear activation function is used in the final layer, leaky rectified linear activation is exploited in all other layers. Even though YOLO is fast and sensitive to false positives, it still has problems with localization and recall.

YOLOv2 \citep{Redmon2016YOLO9000} aims to improve shortcomings of the first version of YOLO by simplifying the network with better representations. The first modification is adding batch normalization which also allows to remove dropout from the model. YOLOv2's classification network is trained on \textit{448 x 448} resolution, instead of \textit{224 x 224.} The network is then shrunk to \textit{416 x 416} in order to obtain locations in the feature map as an odd number. The fully connected layers are removed in YOLOv2 and anchor boxes are added using dimension clusters to predict bounding boxes. Anchor boxes requires manually determined box dimensions. To overcome this problem, k-means clustering is used to determine the bounding box priors. Even though use of anchor boxes results in slight decrease in accuracy, an intermediate increase is obtained in recall. The most significant improvement in YOLOv2 is the new backbone. Instead of using GoogLeNet based architecture, Darknet-19 is proposed which consists of 19 convolutional and 5 maximum pooling layers.

YOLOv3 \citep{Redmon2018YOLOv3} uses logistic regression to calculate objectness score for each predicted bounding box. The bounding boxes are predicted at three different scales by extracting features in all scales using a modified feature pyramid network \citep{Lin2016Feature}. In order to extract more semantic information, features from the earlier layers of the network are concatenated with upsampled features from layers at the later stages. The feature extraction network Darknet-19 is more extended with \textit{3 x 3} and \textit{1 x 1} convolutional layers and skip connections. The improved version has 53 convolutional layer and referred as Darknet-53. Although YOLOv3 performs better with small objects compared to older YOLO versions, this is not the case for medium and large objects.

YOLOv4 \citep{Bochkovskiy2020YOLOv4} aims to reach optimum balance between resolution, layers and parameters in order to obtain accurate results rapidly. Darknet-53 backbone network is improved with Cross Stage Partial (CSP) module \citep{9150780} which is called CSPDarknet-53. In this scenario, the feature map of the base layer is partitioned and merged using a cross-stage hierarchy which provides more gradient flow. Additionally, Spatial pyramid pooling (SPP) \citep{He2014Spatial} is integrated into the backbone of the network (CSPDarknet53 with Mish activation) in order to increase the receptive field. This helps to extract the main characteristic features without slowing the network. Path aggregation network (PANet) with Leaky activation is used for feature extraction. Additionally, some CNN components have been integrated to backbone and detector in order to improve the network further such as dropblock regularization, cross mini-batch normalization, CutMix and Mosaic data augmentation \citep{Bochkovskiy2020YOLOv4}.


\subsection{Faster R-CNN}\label{a65c0108}

Regions with CNN features (R-CNN) \citep{6909475} combines region proposal with high-capacity CNNs that allows to bottom-up region proposals for better localization and performance. R-CNN consists of three parts. The first part creates region proposals using selective search which are not dependent on the class. The first part of the network creates around 2000 region proposals. The second part performs feature extraction using Caffe \citep{Jia2014Caffe} with shared parameters for all classes. Since proposed regions can be in any size, they are dilated and warped to \textit{227 x 227.} Caffe network extracts a fixed size low-dimensional feature vector for each proposed region using five convolutional and two fully connected layers. The final part of the network scores each extracted feature vectors utilizing class-specific linear SVMs. All regions with scores are then analyzed with non-maximum suppression for each class to obtain the best region proposal with the highest IoU.

Feature extraction for each region proposal rather than a whole image increases computational cost and storage space. In order to overcome drawbacks, improve speed and increase accuracy of multi-stage structured R-CNN method, a single stage Fast-RCNN method that jointly learns classification of proposal and refined localization \citep{Girshick2015Fast}. Fast R-CNN network processes the whole image to create a feature map. A region of interest (RoI) pooling layer similar to SPPnet [34] takes the feature map and extracts a fixed size feature vector using maximum pooling applied on each channel. Softmax probability estimates and bounding box positions are then produced by feeding each feature vector into a series of fully connected layers. During training of Fast R-CNN, stochastic gradient descent mini-batches are sampled hierarchically. Computation and memory costs are shared for RoIs created in the same images during forward-backward passes. Additionally, the network jointly optimizes the classifier and the regressor in a single stage.

Faster R-CNN \citep{NIPS2015_14bfa6bb} is based on a region proposal network (RPN). The detector in the network works with the rectangular object proposals by the RPN. Object proposals are then used by Fast R-CNN for detection. RPN also shares created features with Fast R-CNN thus, it does not increase computational cost. RPN also utilizes attention mechanisms that directs Fast R-CNN detection network where to look. The anchors are created for each location as translation invariant and multi-scale. A multi-task loss is calculated by considering log loss and regression loss. Log loss is the classification loss between classes. For a single class airplane detection network, the classification loss calculates loss over airplane versus not plane. The regression loss is calculated once the anchor contains the desired object. Note that an object will have a defined maximum number of anchors. However, the anchor with the minimum loss will be attained as detection \citep{NIPS2015_14bfa6bb}.

\subsection{Accuracy Metrics}\label{yqrOeC0drX}

The results are assessed using Microsoft COCO \citep{Lin2014Microsoft} evaluation metrics which consists of various Average Precision (AP) and Average Recall (AR) values. In this study, we have used the first 3 evaluation metrics of Microsoft COCO to assess test results. These are namely mean average precision (mAP), mAP at 50\% Intersection over Union (IoU) and mAP at 75\% IoU. AP is a summarization metric derived from precision-recall curve. It is calculated by weighted mean of precision values for different recall threshold values varying from 0 to 1:

\begin{equation}
AP = \sum_{n=1} (R_n-R_{n-1})\times P
\end{equation}

True Positives (TPs) are determined by IoU thresholds. For example, an IoU threshold of 50\% means that predicted bounding box will be counted as TP once it has equal or greater than IoU value of 50\% with the ground truth and AP is calculated based on this assumption which is the AP at 50\% IoU threshold and referred as PASCAL VOC metric \citep{everingham2010pascal}. It is the same case for the AP at 75\% IoU threshold which is more strict. mAP is calculated by averaging APs calculated for 10 IoU thresholds from 50\% to 95\% with 0.05 increment for all classes. Since we have only one class in this study, mAP and AP values are identical.


\section{Results and Discussion}\label{a79e4061}

Experiments were conducted in an Intel Core i9-9900K 3.6 GHz CPU and a NVIDIA GeForce RTX 2080 Ti GPU. The training process for YOLOv4 was carried out in Darknet framework. We conducted several initial experiments to find out the best hyperparameter configuration. Our results demonstrated that increasing batch size and subdivision affect the performance positively. This is similar for input image size as well. Higher image sizes provide better performance; however, computing load also increased for bigger image sizes which is directly determined by the GPU. After hyper-tuning experiments, the best configuration for our hardware was found as; the input image of 416 x 416 pixels for the network size, 64 as the batch size and 32 for the subdivision value. The learning rate, decay and momentum were input as 0.001, 0.0005 and 0.949, respectively. Complete-IoU loss was used as the loss function. Finally, different data augmentation methods were implemented but our results show that using only mosaic augmentation improved the results considerably.

Faster R-CNN network were trained on TensorFlow Object Detection API. The input image size was 1024 x 600. Momentum optimizer were utilized with 0.0001 learning rate. We have used random horizontal flip method for data augmentation. Both networks were trained using pre-trained weights from MS COCO dataset. As explained for YOLOv4 experiments, Faster R-CNN hyperparameters for training have been determined empirically. YOLOv4 training took approximately 12 hours to complete while Faster R-CNN training process was around 10 hours.

Deep neural networks have been evaluated using the same test dataset for two different models. The evaluation results are given in Table~\ref{t4Gyr1pH5R}. The evaluation results show that both networks perform well up to 75\% IoU threshold; mAP value of YOLOv4 is 73.02 \%; whereas Faster R-CNN provided slightly better performance with 76.40\%. Although YOLOv4 produces very high 99.15\% value for IoU of 50\%, this value reduces with increasing IoU values and reached to 91.82 \% at IoU of 75\%. YOLOv4 seems superior considering 50\% and 75\% IoU threshold. The decrease rate of AP with increasing IoU is higher for YOLOv4 compared to Faster R-CNN. This indicates that YOLOv4 cannot perform efficiently in higher IoU threshold levels higher than 80\% in our dataset.

\begin{table}[!htbp]
  \centering
  \caption{
    Evaluation results based on average precision
  }
  \label{t4Gyr1pH5R}

  \adjustbox{max width=\textwidth}{%
  \begin{tabular}{*{3}{c}}
    \hline
     & \textbf{YOLOv4} & \textbf{Faster R-CNN} \\
    \hline
    \textbf{mAP} & 73.02\% & 76.40\% \\
    \textbf{mAP@IoU=50\%} & 99.15\% & 96.80\% \\
    \textbf{mAP@IoU=75\%} & 91.82\% & 90.00\% \\
    \hline
  \end{tabular}}
\end{table}

\begin{figure}[!htbp]
  \centering
  \includegraphics[width=1\linewidth]{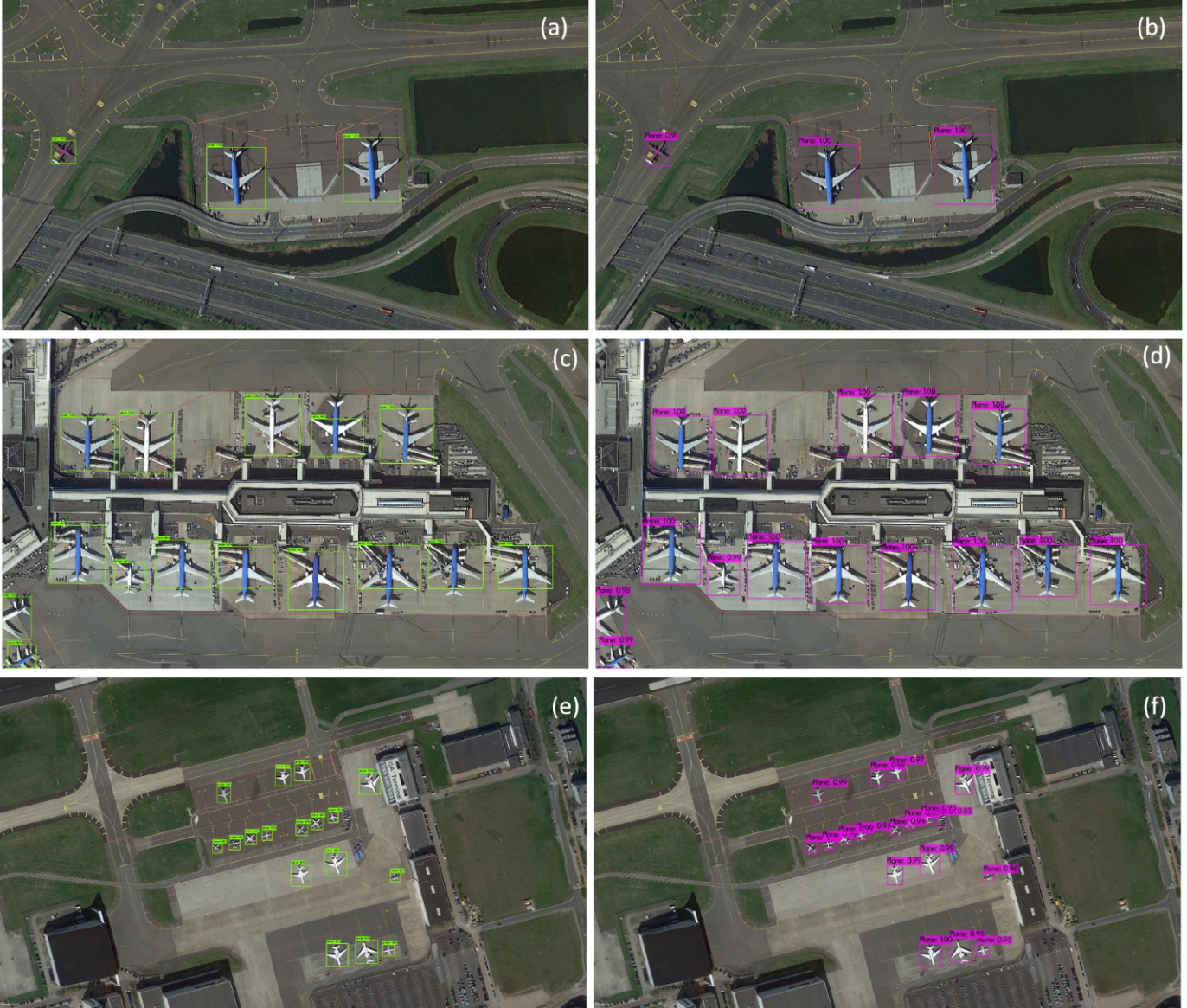}

  \caption{
    Prediction results of Faster R-CNN (green) and YOLOv4 (purple) from Amsterdam Schipol Airport
  }
  \label{XoUMex1Wfm}
\end{figure}

We present some figures to illustrate the obtained results with two different models; where purple boxes represent YOLOv4 and green boxes represent Faster R-CNN results. Since images are collected from different satellites, the real-world coverage of the images is different with respect to spatial resolution while represented with \textit{4800 x 2703} pixels. For very high-resolution images, the airplanes can be seen as very big objects within the scene with respect to image patch size. We will be using the term large scale for these examples such as Figure 2a, b, c and d.~ For high resolution images, the image patch is covering bigger area in which we will be using the term small scale such as Figure 2e and f. The assessment of Figure~\ref{XoUMex1Wfm} including samples from Amsterdam Schiphol Airport shows that airplanes represented with bigger object boxes could be clearly identified with both YOLOv4 and Faster R-CNN architectures as can be seen Figure 2a, b, c and d. However, boundaries of bounding boxes for YOLOv4 seem better at this scale specifically in Figure 2d compared to Faster R-CNN of Figure 2c in which tails of some airplanes are not included within the bounding boxes. We have a small-scale image example in Figure 2e and f, in which there are also airplanes of different sizes are available. Faster R-CNN (Figure 2e) produces better results compared to YOLOv4 specifically for small airplanes.

\begin{figure}[!htbp]
  \centering
  \includegraphics[width=1\linewidth]{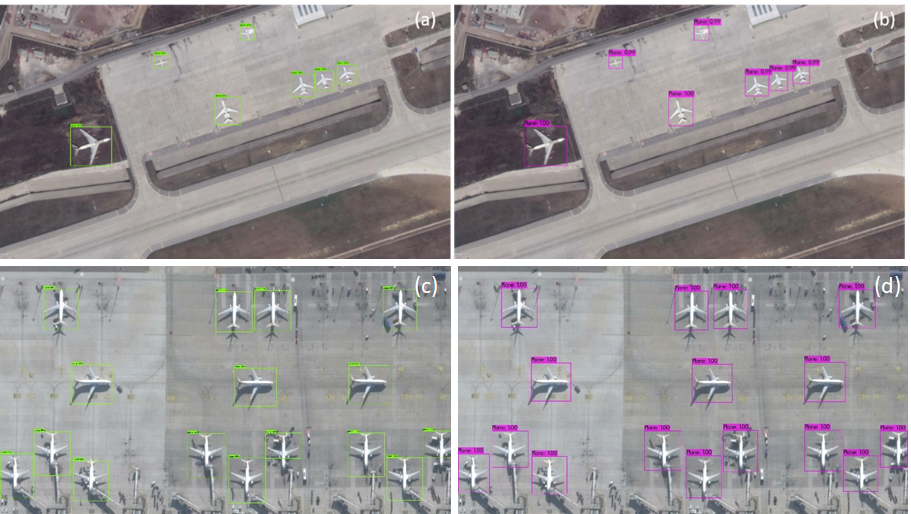}

  \caption{
    Prediction results of Faster R-CNN (green) and YOLOv4 (purple) from Istanbul Sabiha Gokcen Airport
  }
  \label{vR5Sl6z5Zr}
\end{figure}

Figure~\ref{vR5Sl6z5Zr} shows predictions results for a small scale imagery from Istanbul Sabiha Gokcen Airport. On the contrary to Figure 2f, YOLOv4 seem to be performing better in this example (Figure 3b). Both architecture have also detected the small propeller aircraft which is located in the upper left of the image (Figure 3a and b). Figure 3c and d present predictions for commercial planes in large scale in the same airport. In this example, YOLOv4 seems to create better bounding boxes (Figure 3c) than Faster R-CNN (Figure 3d). This may be resulted due to boarding bridges near the planes.

\begin{figure}[!htbp]
  \centering
  \includegraphics[width=1\linewidth]{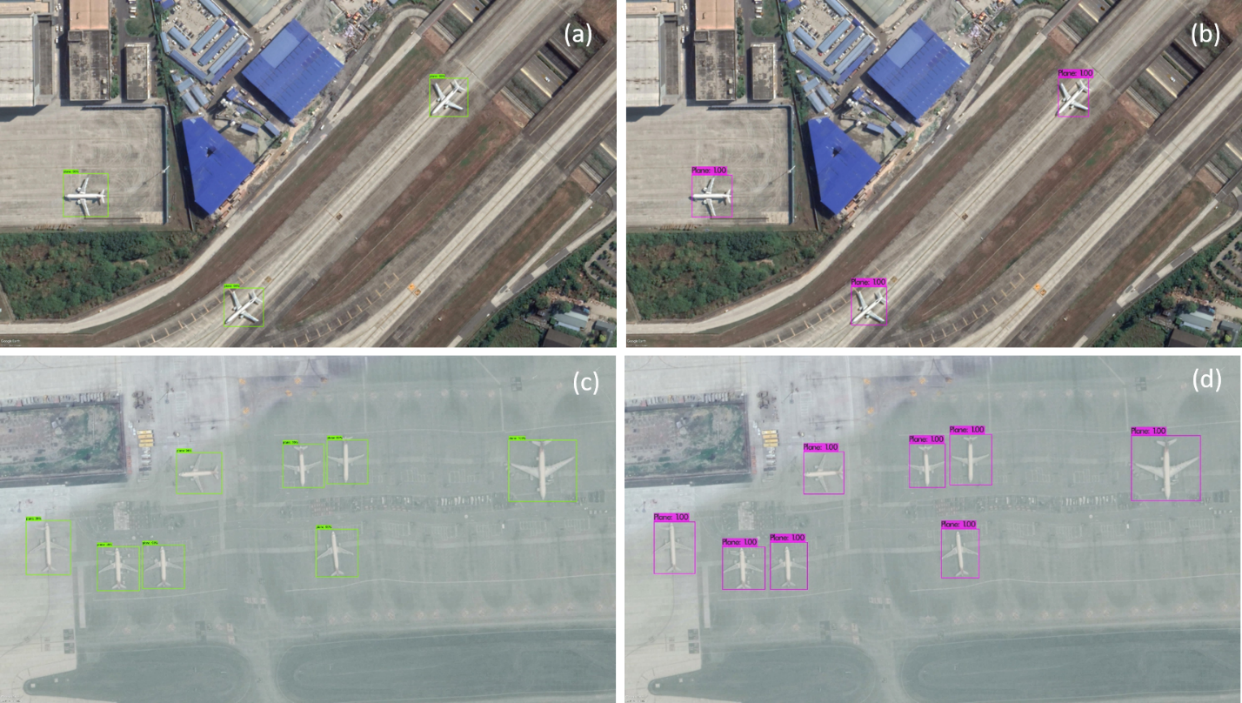}

  \caption{
    Prediction results of YOLOv4 (purple) and Faster R-CNN (green) from Chengdu Shuangliu International
  }
  \label{wZtOovw3pr}
\end{figure}

Our HRPlanes dataset consists of imageries from airport around different parts of the world. Some prediction results from Chengdu Shuangliu International Airport are shown in Figure~\ref{wZtOovw3pr}. Both methods performed sufficiently with similar bounding boxes under clear (Figure 4a and b) and hazy (Figure 4c and d) atmospheric conditions.

\begin{figure}[!htbp]
  \centering
  \includegraphics[width=1\linewidth]{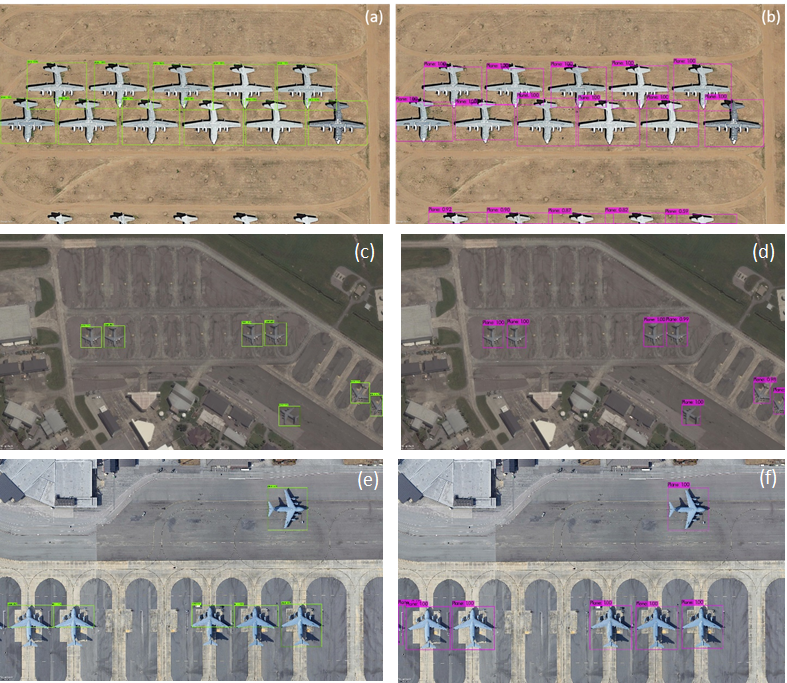}

  \caption{
    Prediction results of Faster R-CNN (green) and YOLOv4 (purple) from Davis Monthan for millitary planes.
  }
  \label{ncMSRxCzaH}
\end{figure}

Military plane samples from Davis Monthan Air Force Base are shown in Figure~\ref{ncMSRxCzaH}. Both prediction results seem similar for fully visible airplanes in terms of both detection and bounding boxes as can be seen on the center part of the image patch in Figure 5a and 5b; whereas, only YOLOv4 is able to detect tails of aircrafts on the southern part of the image patch (Figure 5b). Faster R-CNN could not able to capture parts of the aircrafts in this example. In another military planes example (Figure 5c and d), both architectures performed well and successfully detected seven military planes within the image patch. However, in some rare cases Faster R-CNN does not seem to create accurate bounding boxes for military planes and generate bounding boces representing almost half of the airplanes (Figure 5e). Bounding boces of four out of six airplanes are not completely generated. YOLOv4 could detect all six airplanes and generate bounding boces for all of them for the image patch (Figure 5f).

\begin{figure}[!htbp]
  \centering
  \includegraphics[width=1\linewidth]{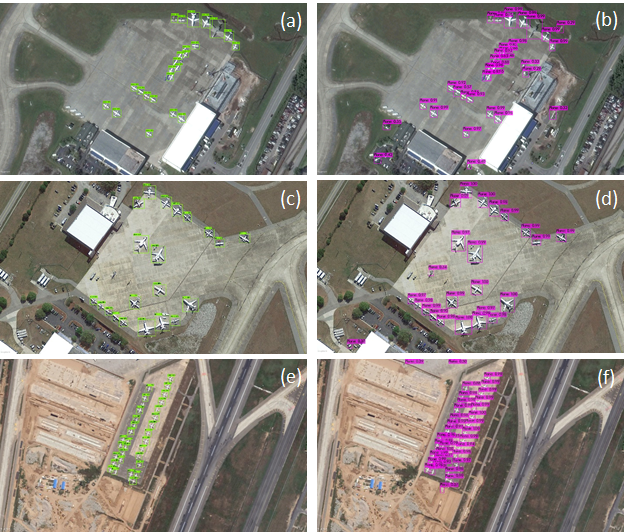}

  \caption{
    Prediction results of Faster R-CNN (green) and YOLOv4 (purple) from a small-scale image with small propeller aircrafts
  }
  \label{wIIw00pOAM}
\end{figure}

Some small scale examples which mostly consists of small propeller aircrafts is given in Figure~\ref{wIIw00pOAM}. Faster R-CNN seems to have significantly performed better in this imageries considering YOLOv4 have created various False Positives (FP) in different parts of the image (Figure 6b, d and f). Some FP detections have lower confidence levels, which could be eliminated by increasing confidence level to 0.50 such as cases in Figure 6b to improve the results of YOLOv4.

\begin{figure}[!htbp]
  \centering
  \includegraphics[width=1\linewidth]{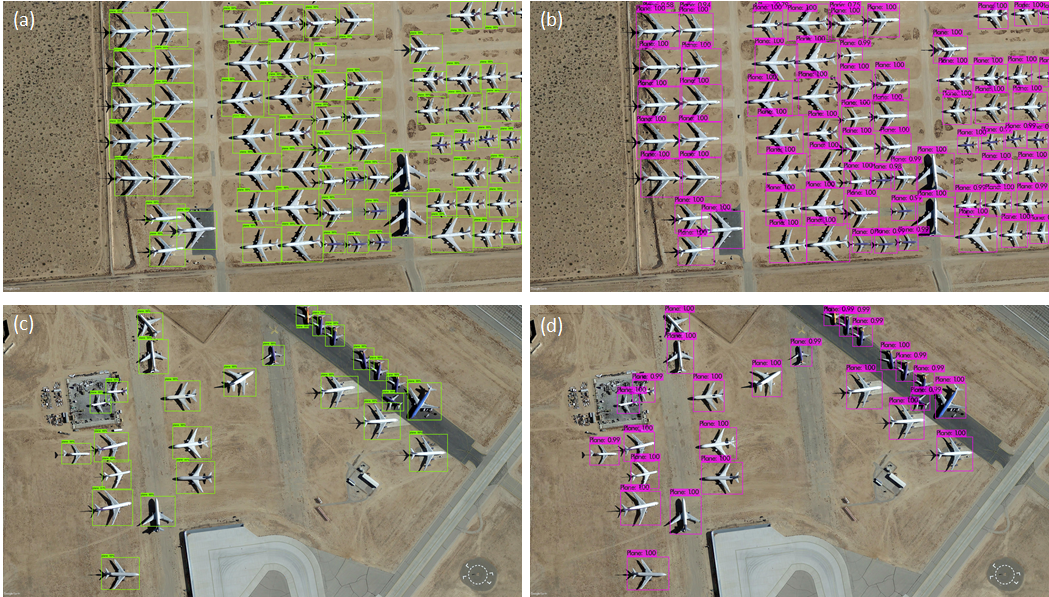}

  \caption{
    Prediction results of Faster R-CNN (green) and YOLOv4 (purple) from a small-scale image with small propeller aircrafts
  }
  \label{bu4SKfxGCZ}
\end{figure}

We show a plane graveyard example in Figure~\ref{bu4SKfxGCZ}. Since the background is not complex and the planes are well aligned, both methods have performed efficiently in both examples. Even in small scale image, YOLOv4 was able to detect plane even though only their nose cones are visible which can be seen in upper right corner of Figure 7b.

\begin{figure}[!htbp]
  \centering
  \includegraphics[width=1\linewidth]{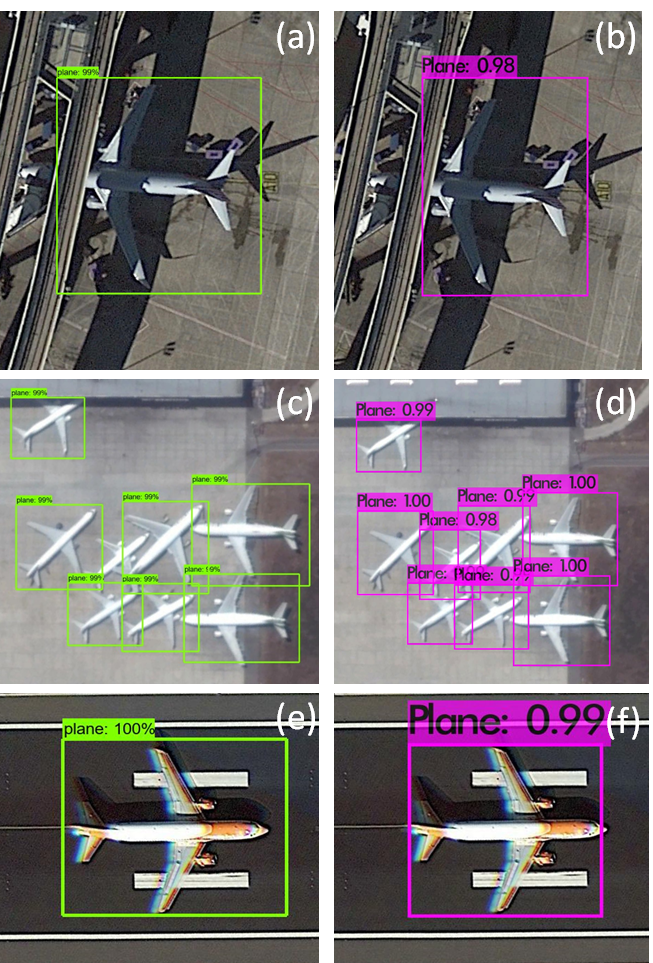}

  \caption{
    Visibility, density and motion blur examples
  }
  \label{IL1Zu5pOWj}
\end{figure}

According to the prediction results, partly absent airplanes are generally detected by YOLOv4 (Figure 5b \& Figure 7b). Only in some rare cases Faster R-CNN is also able to detect them (Figure~\ref{IL1Zu5pOWj}). Additionally, it can be said that both methods can successfully identify airplanes in crowded scenes thanks to non-max suppression technique, even though Faster-RCNN have skipped one plane in Fig 8c. Both networks were even able to detect moving planes which has motion blur effect in the image (Fig 8e and f).

\section{Conclusion}\label{a54d0b7a}

We created a novel airplane detection dataset, called HRPlanes, that includes image patches of commercial and military airplanes. We generated this new dataset to be benchmark for deep learning-based object detection methods specifically for airplane cases. We evaluated performance of YOLOv4 and Faster R-CNN on the created dataset with various experiments with different hyperparameters. Both models have provided satisfactory results at 75\% IoU threshold above 90\% mAP which is quite high. Our results show that the dataset provides highly accurate information to train deep neural networks efficiently. Our proposed models and hyperparameter setting could be used for various remote sensing-based airplane detection tasks and our model weights could be directly used for the inference of similar datasets and transfer learning of different datasets. Moreover, our test dataset could be used by different researchers to evaluate the new model proposals. After further analysis and quality checks, it is planned to share the all dataset publicly. Train and validation sets of the dataset are available upon request from the corresponding author. Currently, the weights of both networks and test dataset are available on: \textit{https://github.com/TolgaBkm/HRPlanes}

\section*{Acknowledgments}

\subsection*{Acknowledgements}

The authors would like to thank all researchers from Istanbul Technical University, Center for Satellite Communications and Remote Sensing for their assistance during data processing. We are also grateful to Google Earth for providing high resolution satellite imagery.

\bibliographystyle{unsrtnat}
\bibliography{main}

\begin{thebibliography}{40}
\providecommand{\natexlab}[1]{#1}
\providecommand{\url}[1]{\texttt{#1}}
\expandafter\ifx\csname urlstyle\endcsname\relax
  \providecommand{\doi}[1]{doi: #1}\else
  \providecommand{\doi}{doi: \begingroup \urlstyle{rm}\Url}\fi

\bibitem[Li et~al.(2020)Li, Wan, Cheng, Meng, and Han]{LI2020296}
Ke~Li, Gang Wan, Gong Cheng, Liqiu Meng, and Junwei Han.
\newblock Object detection in optical remote sensing images: A survey and a new
  benchmark.
\newblock \emph{ISPRS Journal of Photogrammetry and Remote Sensing},
  159:\penalty0 296--307, 2020.
\newblock ISSN 0924-2716.
\newblock \doi{10.1016/j.isprsjprs.2019.11.023}.
\newblock URL
  \url{https://www.sciencedirect.com/science/article/pii/S0924271619302825}.

\bibitem[Alganci et~al.(2020)Alganci, Soydas, and Sertel]{rs12030458}
Ugur Alganci, Mehmet Soydas, and Elif Sertel.
\newblock Comparative {Research} on {Deep} {Learning} {Approaches} for
  {Airplane} {Detection} from {Very} {High}-{Resolution} {Satellite} {Images}.
\newblock \emph{Remote Sensing}, 12\penalty0 (3), 2020.
\newblock ISSN 2072-4292.
\newblock \doi{10.3390/rs12030458}.
\newblock URL \url{https://www.mdpi.com/2072-4292/12/3/458}.

\bibitem[Liu et~al.(2013)Liu, Sun, Fu, and Wang]{6353895}
Ge~Liu, Xian Sun, Kun Fu, and Hongqi Wang.
\newblock Aircraft {Recognition} in {High}-{Resolution} {Satellite} {Images}
  {Using} {Coarse}-to-{Fine} {Shape} {Prior}.
\newblock \emph{IEEE Geoscience and Remote Sensing Letters}, 10\penalty0
  (3):\penalty0 573--577, 2013.
\newblock \doi{10.1109/LGRS.2012.2214022}.

\bibitem[Xu and Duan(2010)]{XU20101759}
Chunfan Xu and Haibin Duan.
\newblock Artificial bee colony ({ABC}) optimized edge potential function
  ({EPF}) approach to target recognition for low-altitude aircraft.
\newblock \emph{Pattern Recognition Letters}, 31\penalty0 (13):\penalty0
  1759--1772, 2010.
\newblock ISSN 0167-8655.
\newblock \doi{10.1016/j.patrec.2009.11.018}.
\newblock URL
  \url{https://www.sciencedirect.com/science/article/pii/S0167865509003298}.
\newblock Meta-heuristic Intelligence Based Image Processing.

\bibitem[Cheng and Han(2016)]{CHENG201611}
Gong Cheng and Junwei Han.
\newblock A survey on object detection in optical remote sensing images.
\newblock \emph{ISPRS Journal of Photogrammetry and Remote Sensing},
  117:\penalty0 11--28, 2016.
\newblock ISSN 0924-2716.
\newblock \doi{10.1016/j.isprsjprs.2016.03.014}.
\newblock URL
  \url{https://www.sciencedirect.com/science/article/pii/S0924271616300144}.

\bibitem[Sun et~al.(2012)Sun, Sun, Wang, Li, and Li]{5982082}
Hao Sun, Xian Sun, Hongqi Wang, Yu~Li, and Xiangjuan Li.
\newblock Automatic {Target} {Detection} in {High}-{Resolution} {Remote}
  {Sensing} {Images} {Using} {Spatial} {Sparse} {Coding} {Bag}-of-{Words}
  {Model}.
\newblock \emph{IEEE Geoscience and Remote Sensing Letters}, 9\penalty0
  (1):\penalty0 109--113, 2012.
\newblock \doi{10.1109/LGRS.2011.2161569}.

\bibitem[Zhang et~al.(2014)Zhang, Sun, Fu, Wang, and Wang]{6512596}
Wanceng Zhang, Xian Sun, Kun Fu, Chenyuan Wang, and Hongqi Wang.
\newblock Object {Detection} in {High}-{Resolution} {Remote} {Sensing} {Images}
  {Using} {Rotation} {Invariant} {Parts} {Based} {Model}.
\newblock \emph{IEEE Geoscience and Remote Sensing Letters}, 11\penalty0
  (1):\penalty0 74--78, 2014.
\newblock \doi{10.1109/LGRS.2013.2246538}.

\bibitem[Zhang et~al.(2015)Zhang, Sun, Wang, and Fu]{ZHANG201530}
Wanceng Zhang, Xian Sun, Hongqi Wang, and Kun Fu.
\newblock A generic discriminative part-based model for geospatial object
  detection in optical remote sensing images.
\newblock \emph{ISPRS Journal of Photogrammetry and Remote Sensing},
  99:\penalty0 30--44, 2015.
\newblock ISSN 0924-2716.
\newblock \doi{10.1016/j.isprsjprs.2014.10.007}.
\newblock URL
  \url{https://www.sciencedirect.com/science/article/pii/S0924271614002573}.

\bibitem[Lei et~al.(2012)Lei, Fang, Huo, and Li]{6043875}
Zhen Lei, Tao Fang, Hong Huo, and Deren Li.
\newblock Rotation-{Invariant} {Object} {Detection} of {Remotely} {Sensed}
  {Images} {Based} on {Texton} {Forest} and {Hough} {Voting}.
\newblock \emph{IEEE Transactions on Geoscience and Remote Sensing},
  50\penalty0 (4):\penalty0 1206--1217, 2012.
\newblock \doi{10.1109/TGRS.2011.2166966}.

\bibitem[Liu and Shi(2014)]{LIU20145327}
Liu Liu and Zhenwei Shi.
\newblock Airplane detection based on rotation invariant and sparse coding in
  remote sensing images.
\newblock \emph{Optik}, 125\penalty0 (18):\penalty0 5327--5333, 2014.
\newblock ISSN 0030-4026.
\newblock \doi{10.1016/j.ijleo.2014.06.062}.
\newblock URL
  \url{https://www.sciencedirect.com/science/article/pii/S0030402614007074}.

\bibitem[Ball et~al.(2017)Ball, Anderson, and Sr.]{Ball2017Comprehensive}
John~E. Ball, Derek~T. Anderson, and Chee Seng~Chan Sr.
\newblock Comprehensive survey of deep learning in remote sensing: theories,
  tools, and challenges for the community.
\newblock \emph{Journal of Applied Remote Sensing}, 11\penalty0 (4):\penalty0 1
  -- 54, 2017.
\newblock \doi{10.1117/1.JRS.11.042609}.
\newblock URL \url{https://doi.org/10.1117/1.JRS.11.042609}.

\bibitem[Chen et~al.(2018)Chen, Zhang, and Ouyang]{rs10010139}
Zhong Chen, Ting Zhang, and Chao Ouyang.
\newblock End-to-{End} {Airplane} {Detection} {Using} {Transfer} {Learning} in
  {Remote} {Sensing} {Images}.
\newblock \emph{Remote Sensing}, 10\penalty0 (1), 2018.
\newblock ISSN 2072-4292.
\newblock \doi{10.3390/rs10010139}.
\newblock URL \url{https://www.mdpi.com/2072-4292/10/1/139}.

\bibitem[Xu et~al.(2018)Xu, Zhu, Xin, Li, Qi, and Ma]{s18072335}
Yuelei Xu, Mingming Zhu, Peng Xin, Shuai Li, Min Qi, and Shiping Ma.
\newblock Rapid {Airplane} {Detection} in {Remote} {Sensing} {Images} {Based}
  on {Multilayer} {Feature} {Fusion} in {Fully} {Convolutional} {Neural}
  {Networks}.
\newblock \emph{Sensors}, 18\penalty0 (7), 2018.
\newblock ISSN 1424-8220.
\newblock \doi{10.3390/s18072335}.
\newblock URL \url{https://www.mdpi.com/1424-8220/18/7/2335}.

\bibitem[Zhu et~al.(2019)Zhu, Xu, Ma, Li, Ma, and Han]{rs11091062}
Mingming Zhu, Yuelei Xu, Shiping Ma, Shuai Li, Hongqiang Ma, and Yongsai Han.
\newblock Effective {Airplane} {Detection} in {Remote} {Sensing} {Images}
  {Based} on {Multilayer} {Feature} {Fusion} and {Improved} {Nonmaximal}
  {Suppression} {Algorithm}.
\newblock \emph{Remote Sensing}, 11\penalty0 (9), 2019.
\newblock ISSN 2072-4292.
\newblock \doi{10.3390/rs11091062}.
\newblock URL \url{https://www.mdpi.com/2072-4292/11/9/1062}.

\bibitem[Wu et~al.(2020)Wu, Weise, Wang, and Wang]{9178761}
Zhi-Ze Wu, Thomas Weise, Yan Wang, and Yongjun Wang.
\newblock Convolutional {Neural} {Network} {Based} {Weakly} {Supervised}
  {Learning} for {Aircraft} {Detection} {From} {Remote} {Sensing} {Image}.
\newblock \emph{IEEE Access}, 8:\penalty0 158097--158106, 2020.
\newblock \doi{10.1109/ACCESS.2020.3019956}.

\bibitem[Zhou et~al.(2021)Zhou, Yan, Shan, Zheng, Liu, Zuo, and
  Qiao]{zhou2021aircraft}
Liming Zhou, Haoxin Yan, Yingzi Shan, Chang Zheng, Yang Liu, Xianyu Zuo, and
  Baojun Qiao.
\newblock Aircraft detection for remote sensing images based on deep
  convolutional neural networks.
\newblock \emph{Journal of Electrical and Computer Engineering}, 2021, 2021.

\bibitem[Ji et~al.(2021)Ji, Ming, Zeng, Yu, Qing, Du, and Zhang]{rs13112207}
Fengcheng Ji, Dongping Ming, Beichen Zeng, Jiawei Yu, Yuanzhao Qing, Tongyao
  Du, and Xinyi Zhang.
\newblock Aircraft {Detection} in {High} {Spatial} {Resolution} {Remote}
  {Sensing} {Images} {Combining} {Multi}-{Angle} {Features} {Driven} and
  {Majority} {Voting} {CNN}.
\newblock \emph{Remote Sensing}, 13\penalty0 (11), 2021.
\newblock ISSN 2072-4292.
\newblock \doi{10.3390/rs13112207}.
\newblock URL \url{https://www.mdpi.com/2072-4292/13/11/2207}.

\bibitem[Shi et~al.(2021)Shi, Tang, Wang, Xu, Liu, and Zhang]{Shi2021Aircraft}
Lukui Shi, Zhenjie Tang, Tiantian Wang, Xia Xu, Jing Liu, and Jun Zhang.
\newblock Aircraft detection in remote sensing images based on deconvolution
  and position attention.
\newblock \emph{International Journal of Remote Sensing}, 42\penalty0
  (11):\penalty0 4241--4260, 2021.
\newblock \doi{10.1080/01431161.2021.1892858}.
\newblock URL
  \url{%20%0A%20%20%20%20%20%20%20%20https://doi.org/10.1080/01431161.2021.1892858%0A%20%20%20%20%0A}.

\bibitem[Wu et~al.(2021)Wu, Feng, Cao, Zeng, Feng, Wu, and Huang]{s21082618}
Qifan Wu, Daqiang Feng, Changqing Cao, Xiaodong Zeng, Zhejun Feng, Jin Wu, and
  Ziqiang Huang.
\newblock Improved {Mask} {R}-{CNN} for {Aircraft} {Detection} in {Remote}
  {Sensing} {Images}.
\newblock \emph{Sensors}, 21\penalty0 (8), 2021.
\newblock ISSN 1424-8220.
\newblock \doi{10.3390/s21082618}.
\newblock URL \url{https://www.mdpi.com/1424-8220/21/8/2618}.

\bibitem[Zeng et~al.(2022)Zeng, Ming, Ji, Yu, Xu, Zhang, and Lian]{Zeng2022Top}
Beichen Zeng, Dongping Ming, Fengcheng Ji, Jiawei Yu, Lu~Xu, Liang Zhang, and
  Xinyi Lian.
\newblock Top-{Down} aircraft detection in large-scale scenes based on
  multi-source data and {FEF}-{R}-{CNN}.
\newblock \emph{International Journal of Remote Sensing}, 43\penalty0
  (3):\penalty0 1108--1130, 2022.
\newblock \doi{10.1080/01431161.2022.2034194}.
\newblock URL
  \url{%20%0A%20%20%20%20%20%20%20%20https://doi.org/10.1080/01431161.2022.2034194%0A%20%20%20%20%0A}.

\bibitem[Chen et~al.(2022)Chen, Liu, Xu, Xie, Zuo, and Cao]{s22010319}
Xin Chen, Jinghong Liu, Fang Xu, Zhihua Xie, Yujia Zuo, and Lihua Cao.
\newblock A {Novel} {Method} of {Aircraft} {Detection} under {Complex}
  {Background} {Based} on {Circular} {Intensity} {Filter} and {Rotation}
  {Invariant} {Feature}.
\newblock \emph{Sensors}, 22\penalty0 (1), 2022.
\newblock ISSN 1424-8220.
\newblock \doi{10.3390/s22010319}.
\newblock URL \url{https://www.mdpi.com/1424-8220/22/1/319}.

\bibitem[Xia et~al.(2017)Xia, Bai, Ding, Zhu, Belongie, Luo, Datcu, Pelillo,
  and Zhang]{Xia2017DOTA}
Gui-Song Xia, Xiang Bai, Jian Ding, Zhen Zhu, Serge Belongie, Jiebo Luo, Mihai
  Datcu, Marcello Pelillo, and Liangpei Zhang.
\newblock Dota: A {Large}-scale {Dataset} for {Object} {Detection} in {Aerial}
  {Images}.
\newblock 2017.
\newblock \doi{10.48550/ARXIV.1711.10398}.
\newblock URL \url{https://arxiv.org/abs/1711.10398}.

\bibitem[Zamir et~al.(2019)Zamir, Arora, Gupta, Khan, Sun, Khan, Zhu, Shao,
  Xia, and Bai]{Zamir2019iSAID}
Syed~Waqas Zamir, Aditya Arora, Akshita Gupta, Salman Khan, Guolei Sun,
  Fahad~Shahbaz Khan, Fan Zhu, Ling Shao, Gui-Song Xia, and Xiang Bai.
\newblock \emph{iSAID: A {Large}-scale {Dataset} for {Instance} {Segmentation}
  in {Aerial} {Images}}.
\newblock arXiv, 2019.
\newblock \doi{10.48550/ARXIV.1905.12886}.
\newblock URL \url{https://arxiv.org/abs/1905.12886}.

\bibitem[Lam et~al.(2018)Lam, Kuzma, McGee, Dooley, Laielli, Klaric, Bulatov,
  and McCord]{Lam2018xView}
Darius Lam, Richard Kuzma, Kevin McGee, Samuel Dooley, Michael Laielli, Matthew
  Klaric, Yaroslav Bulatov, and Brendan McCord.
\newblock \emph{xView: Objects in {Context} in {Overhead} {Imagery}}.
\newblock arXiv, 2018.
\newblock \doi{10.48550/ARXIV.1802.07856}.
\newblock URL \url{https://arxiv.org/abs/1802.07856}.

\bibitem[Shermeyer et~al.(2020)Shermeyer, Hossler, Van~Etten, Hogan, Lewis, and
  Kim]{Shermeyer2020RarePlanes}
Jacob Shermeyer, Thomas Hossler, Adam Van~Etten, Daniel Hogan, Ryan Lewis, and
  Daeil Kim.
\newblock \emph{RarePlanes: Synthetic {Data} {Takes} {Flight}}.
\newblock arXiv, 2020.
\newblock \doi{10.48550/ARXIV.2006.02963}.
\newblock URL \url{https://arxiv.org/abs/2006.02963}.

\bibitem[{HyperLabel}(2020)]{hyperlabel_2020}
{HyperLabel}.
\newblock \emph{Fast and easy data annotation}.
\newblock 2020.
\newblock URL \url{https://docs.hyperlabel.com/}.

\bibitem[Redmon et~al.(2015)Redmon, Divvala, Girshick, and
  Farhadi]{Redmon2015You}
Joseph Redmon, Santosh Divvala, Ross Girshick, and Ali Farhadi.
\newblock \emph{You {Only} {Look} {Once}: Unified, {Real}-{Time} {Object}
  {Detection}}.
\newblock arXiv, 2015.
\newblock \doi{10.48550/ARXIV.1506.02640}.
\newblock URL \url{https://arxiv.org/abs/1506.02640}.

\bibitem[Szegedy et~al.(2015)Szegedy, Liu, Jia, Sermanet, Reed, Anguelov,
  Erhan, Vanhoucke, and Rabinovich]{7298594}
Christian Szegedy, Wei Liu, Yangqing Jia, Pierre Sermanet, Scott Reed, Dragomir
  Anguelov, Dumitru Erhan, Vincent Vanhoucke, and Andrew Rabinovich.
\newblock Going deeper with convolutions.
\newblock In \emph{2015 {IEEE} {Conference} on {Computer} {Vision} and
  {Pattern} {Recognition} ({CVPR})}, pages 1--9, 2015.
\newblock \doi{10.1109/CVPR.2015.7298594}.

\bibitem[Redmon and Farhadi(2016)]{Redmon2016YOLO9000}
Joseph Redmon and Ali Farhadi.
\newblock \emph{YOLO9000: Better, {Faster}, {Stronger}}.
\newblock arXiv, 2016.
\newblock \doi{10.48550/ARXIV.1612.08242}.
\newblock URL \url{https://arxiv.org/abs/1612.08242}.

\bibitem[Redmon and Farhadi(2018)]{Redmon2018YOLOv3}
Joseph Redmon and Ali Farhadi.
\newblock \emph{YOLOv3: An {Incremental} {Improvement}}.
\newblock arXiv, 2018.
\newblock \doi{10.48550/ARXIV.1804.02767}.
\newblock URL \url{https://arxiv.org/abs/1804.02767}.

\bibitem[Lin et~al.(2016)Lin, Doll{\' a}r, Girshick, He, Hariharan, and
  Belongie]{Lin2016Feature}
Tsung-Yi Lin, Piotr Doll{\' a}r, Ross Girshick, Kaiming He, Bharath Hariharan,
  and Serge Belongie.
\newblock \emph{Feature {Pyramid} {Networks} for {Object} {Detection}}.
\newblock arXiv, 2016.
\newblock \doi{10.48550/ARXIV.1612.03144}.
\newblock URL \url{https://arxiv.org/abs/1612.03144}.

\bibitem[Bochkovskiy et~al.(2020)Bochkovskiy, Wang, and
  Liao]{Bochkovskiy2020YOLOv4}
Alexey Bochkovskiy, Chien-Yao Wang, and Hong-Yuan~Mark Liao.
\newblock \emph{YOLOv4: Optimal {Speed} and {Accuracy} of {Object}
  {Detection}}.
\newblock arXiv, 2020.
\newblock \doi{10.48550/ARXIV.2004.10934}.
\newblock URL \url{https://arxiv.org/abs/2004.10934}.

\bibitem[Wang et~al.(2020)Wang, Mark~Liao, Wu, Chen, Hsieh, and Yeh]{9150780}
Chien-Yao Wang, Hong-Yuan Mark~Liao, Yueh-Hua Wu, Ping-Yang Chen, Jun-Wei
  Hsieh, and I-Hau Yeh.
\newblock Cspnet: A {New} {Backbone} that can {Enhance} {Learning} {Capability}
  of {CNN}.
\newblock In \emph{2020 {IEEE}/{CVF} {Conference} on {Computer} {Vision} and
  {Pattern} {Recognition} {Workshops} ({CVPRW})}, pages 1571--1580, 2020.
\newblock \doi{10.1109/CVPRW50498.2020.00203}.

\bibitem[He et~al.(2014)He, Zhang, Ren, and Sun]{He2014Spatial}
Kaiming He, Xiangyu Zhang, Shaoqing Ren, and Jian Sun.
\newblock Spatial {Pyramid} {Pooling} in {Deep} {Convolutional} {Networks} for
  {Visual} {Recognition}.
\newblock In David Fleet, Tomas Pajdla, Bernt Schiele, and Tinne Tuytelaars,
  editors, \emph{Computer {Vision} -- {ECCV} 2014}, pages 346--361, Cham, 2014.
  Springer International Publishing.

\bibitem[Girshick et~al.(2014)Girshick, Donahue, Darrell, and Malik]{6909475}
Ross Girshick, Jeff Donahue, Trevor Darrell, and Jitendra Malik.
\newblock Rich {Feature} {Hierarchies} for {Accurate} {Object} {Detection} and
  {Semantic} {Segmentation}.
\newblock In \emph{2014 {IEEE} {Conference} on {Computer} {Vision} and
  {Pattern} {Recognition}}, pages 580--587, 2014.
\newblock \doi{10.1109/CVPR.2014.81}.

\bibitem[Jia et~al.(2014)Jia, Shelhamer, Donahue, Karayev, Long, Girshick,
  Guadarrama, and Darrell]{Jia2014Caffe}
Yangqing Jia, Evan Shelhamer, Jeff Donahue, Sergey Karayev, Jonathan Long, Ross
  Girshick, Sergio Guadarrama, and Trevor Darrell.
\newblock Caffe: Convolutional {Architecture} for {Fast} {Feature} {Embedding}.
\newblock In \emph{Proceedings of the 22nd {ACM} {International} {Conference}
  on {Multimedia}}, MM '14, pages 675--678, Orlando, Florida, USA, 2014.
  Association for Computing Machinery.
\newblock \doi{10.1145/2647868.2654889}.
\newblock URL \url{https://doi.org/10.1145/2647868.2654889}.

\bibitem[Girshick(2015)]{Girshick2015Fast}
Ross Girshick.
\newblock \emph{Fast {R}-{CNN}}.
\newblock arXiv, 2015.
\newblock \doi{10.48550/ARXIV.1504.08083}.
\newblock URL \url{https://arxiv.org/abs/1504.08083}.

\bibitem[Ren et~al.(2015)Ren, He, Girshick, and Sun]{NIPS2015_14bfa6bb}
Shaoqing Ren, Kaiming He, Ross Girshick, and Jian Sun.
\newblock Faster {R}-{CNN}: Towards {Real}-{Time} {Object} {Detection} with
  {Region} {Proposal} {Networks}.
\newblock In C.~Cortes, N.~Lawrence, D.~Lee, M.~Sugiyama, and R.~Garnett,
  editors, \emph{Advances in {Neural} {Information} {Processing} {Systems}},
  volume~28. Curran Associates, Inc., 2015.
\newblock URL
  \url{https://proceedings.neurips.cc/paper/2015/file/14bfa6bb14875e45bba028a21ed38046-Paper.pdf}.

\bibitem[Lin et~al.(2014)Lin, Maire, Belongie, Hays, Perona, Ramanan, Doll{\'
  a}r, and Zitnick]{Lin2014Microsoft}
Tsung-Yi Lin, Michael Maire, Serge Belongie, James Hays, Pietro Perona, Deva
  Ramanan, Piotr Doll{\' a}r, and C.~Lawrence Zitnick.
\newblock Microsoft {COCO}: Common {Objects} in {Context}.
\newblock In David Fleet, Tomas Pajdla, Bernt Schiele, and Tinne Tuytelaars,
  editors, \emph{Computer {Vision} -- {ECCV} 2014}, pages 740--755, Cham, 2014.
  Springer International Publishing.

\bibitem[Everingham et~al.(2010)Everingham, Van~Gool, Williams, Winn, and
  Zisserman]{everingham2010pascal}
Mark Everingham, Luc Van~Gool, Christopher~KI Williams, John Winn, and Andrew
  Zisserman.
\newblock The pascal visual object classes (voc) challenge.
\newblock \emph{International journal of computer vision}, 88\penalty0
  (2):\penalty0 303--338, 2010.

\end{thebibliography}

\end{document}